\pdfoutput=1
\documentclass{article}
\usepackage{blindtext}
\usepackage{jmlr2e}

\usepackage[utf8]{inputenc}
\usepackage[T1]{fontenc}
\usepackage{hyperref}
\usepackage{url}
\usepackage{booktabs}
\usepackage{amsfonts}
\usepackage{nicefrac}
\usepackage{microtype}
\usepackage{xcolor}
\usepackage{amsmath}
\usepackage{listings}

\definecolor{mybgcolor}{RGB}{225,215,255}
\definecolor{stringcolor}{RGB}{208,0,0}
\definecolor{commentcolor}{RGB}{15,127,141}
\usepackage{threeparttable}
\usepackage{wasysym}
\usepackage{subfigure}
\usepackage{caption}

\ShortHeadings{Aequitas Flow}{Jesus, Saleiro, Silva, Jorge, Ribeiro, Gama, Bizarro, Ghani}
\firstpageno{1}

\usepackage{lastpage}
\jmlrheading{25}{2024}{1-\pageref{LastPage}}{5/24; Revised
9/24}{9/24}{24-0677}{Sérgio Jesus, Pedro Saleiro, Inês Oliveira e Silva, Beatriz M. Jorge, Rita P. Ribeiro, João Gama, Pedro Bizarro, Rayid Ghani}
\ShortHeadings{Aequitas Flow: Streamlining Fair ML Experimentation}{Jesus, Saleiro, Silva, Jorge, Ribeiro, Gama, Bizarro, Ghani}

\begin{document}

\title{Aequitas Flow: Streamlining Fair ML Experimentation}

\author{\name Sérgio Jesus\textsuperscript{1,2} \email sergio.jesus@feedzai.com \\
\name Pedro Saleiro\textsuperscript{1} \email pedro.saleiro@feedzai.com\\
\name Inês Oliveira e Silva\textsuperscript{1} \email ines.silva@feedzai.com \\
\name Beatriz M. Jorge\textsuperscript{1} \email beatriz.jorge@feedzai.com\\
\name Rita P. Ribeiro\textsuperscript{2} \email rpribeiro@fc.up.pt\\
\name João Gama\textsuperscript{2} \email jgama@fep.up.pt\\
\name Pedro Bizarro\textsuperscript{1} \email pedro.bizarro@feedzai.com \\
\name Rayid Ghani\textsuperscript{3} \email rayid@cmu.edu\\
\addr \textsuperscript{1}Feedzai \\ 
\textsuperscript{2}University of Porto\\
\textsuperscript{3}Carnegie Mellon University}

\editor{Sebastian Schelter}

\maketitle

\begin{abstract}%

Aequitas Flow is an open-source framework and toolkit for end-to-end Fair Machine Learning (ML) experimentation, and benchmarking in Python. This package fills integration gaps that  exist in other fair ML packages. In addition to the existing audit capabilities in Aequitas, the Aequitas Flow module provides a pipeline for fairness-aware model training, hyperparameter optimization, and evaluation, enabling easy-to-use and rapid experiments and analysis of results. Aimed at ML practitioners and researchers, the framework offers implementations of methods, datasets, metrics, and standard interfaces for these components to improve extensibility. By facilitating the development of fair ML practices, Aequitas Flow hopes to enhance the incorporation of fairness concepts in AI systems making AI systems more robust and fair.

\end{abstract}


\section{Introduction}
\label{sec:introduction}

% We have to reduce this section somehow.
%Machine Learning (ML) fairness is a critical research area, especially in domains such as hiring~\citep{dastin2018}, healthcare~\citep{igoe2021}, criminal justice~\citep{angwin2016,chouldechova2017}, and fraud detection~\citep{zhang2019,bartlett2019,jesus2022}. Numerous studies define metrics and properties of algorithmic fairness~\citep{chouldechova2017,calders2010,dwork2012,feldman2015,hardt2016,davies2017} and propose methods for fairer models~\citep{fish2016,calmon2017,zafar2017,cotter2018}. There are, however, gaps in the user experience and integration of existing tools~\citep{lee2021}, hindering end-to-end experimentation. This, in turn, makes empirical studies challenging, scarce, and often with a small sample of methods, metrics, and datasets~\citep{friedler2019, lamba2021}. The lack of comprehensive studies ultimately affects the practical adoption of fair ML methods, as critical domains require strong evidence before incorporating novel techniques.

Developing Machine Learning (ML) and Artificial Intelligence (AI) systems that result in fairness and equity is a critical topic, especially as such systems get used in high-stakes settings such as hiring~\citep{dastin2018}, healthcare~\citep{igoe2021}, criminal justice~\citep{angwin2016,chouldechova2017}, and financial services ~\citep{zhang2019,bartlett2019,jesus2022}. While numerous studies define metrics and properties of algorithmic fairness~\citep{chouldechova2017,calders2010,dwork2012,feldman2015,hardt2016,davies2017} and propose methods for fairer models~\citep{fish2016,calmon2017,zafar2017,cotter2018}, gaps in the implementation, user experience. and integration of existing tools hinder end-to-end experimentation~\citep{lee2021} and benchmarking. This makes empirical studies and practical use challenging, scarce, and often limited in scope~\citep{friedler2019, lamba2021}, ultimately affecting the adoption of fair ML methods in real-world high-stakes settings.

% Aequitas flow as the new version of Aequitas in the descriptions
%This paper introduces Aequitas Flow, an open-source framework that provides reproducible and extensible end-to-end fair ML experimentation. The package is tailored for researchers to validate novel bias mitigation methods and for practitioners to train fairness-aware models. Aequitas Flow is the latest release of Aequitas\footnote{\url{https://github.com/dssg/aequitas}}~\citep{saleiro2018}, a toolkit for bias auditing and reporting of ML models. A comparison between Aequitas Flow and other fair ML packages, including previous versions of Aequitas, is presented in Table \ref{tab:features}.

This paper introduces Aequitas Flow, an open-source framework for reproducible and extensible end-to-end fair ML experimentation that extends Aequitas, our original bias audit toolkit. The goal is to help 1)researchers compare and benchmark new methods they develop against existing methods in a systematic and reproducible manner and 2) practitioners easily evaluate existing bias mitigation methods and deploy ones that best match their goals.

 Table \ref{tab:features} compares Aequitas Flow, the latest release of the Aequitas package \footnote{\url{https://github.com/dssg/aequitas}}~\citep{saleiro2018} to other fair ML packages to highlight some of the key gaps we aimed to fill with this paper.

\begin{table}[t]
    \centering
    \newcommand*\feature[1]{\ifcase#1 -\or\LEFTcircle\or\CIRCLE\fi}
    \newcommand*\f[4]{\feature#1&\feature#2&\feature#3&\feature#4}
    \makeatletter
    \newcommand*\ex[2]{#1&\f#2}
    \makeatother
    \begin{threeparttable}
    \caption{Comparison of packages for training and evaluation of fair ML Methods.}
    \label{tab:features}
    \begin{tabular}{@{}lcccc}
        \toprule
        & \multicolumn{4}{c}{Packages} \\
        \cline{2-5}
        Functionalities  & AIF360 & Fairlearn & Aequitas & Aequitas Flow\\
        \midrule
        \ex{Group fairness metrics} {1122}\\
        \ex{Pre-processing methods} {2102}\\
        \ex{In-processing methods} {1202}\\
        \ex{Post-processing methods} {2202}\\
        \ex{Standardized interfaces for extensibility} {1102}\\
        \ex{Hyperparameter optimization pipeline} {0002}\\
        \ex{Binary classification}{2222}\\
        \ex{Regression}{2200}\\
        \ex{Model selection}{0012}\\
        \ex{Methods comparison}{0002}\\
        \ex{Plotting methods} {0122}\\
        \bottomrule
    \end{tabular}
    \begin{tablenotes}
        \footnotesize
        \item $\feature2~\text{exists in package}$; $\feature1~\text{partially exists in package}$; $\text{\feature0}~\text{does not exist in package}$.
    \end{tablenotes}    
    \end{threeparttable}
\end{table}

%The two more popular packages for fair ML are Fairlearn~\citep{werts2023}, and AIF360~\citep{bellamy2018}. Both aim to facilitate adoption with an array of methods~\citep{feldman2015,hardt2016,agarwal2018}, common fairness metrics~\citep{hardt2016} and datasets used in the literature~\citep{kohavi1995,angwin2016,dua2019,ding2021}. However, these have specific issues that hinder their usability as standard toolkits for fairness studies in the ML community. The first issue is the lack of a defined pipeline for experimentation. Users must implement or resort to external packages for fundamental tasks of ML experimentation, such as dataset splitting, hyperparameter optimization, thresholding, and results visualization. The second is related to consistency between classes of the same group. Although some classes might appear to operate similarly, they receive different inputs and outputs or operate under different method names. This forces the users to customize the code depending on the method used. An example of this problem is in the \texttt{DisparateImpactRemover} class, in AIF360 where most of the methods of the parent class are not implemented. These two issues create a high entry barrier for the usage of the packages. In this work, we propose to tackle the lack of standardized tools for experimentation with fair ML, with an emphasis on the extensibility of methods, datasets, and metrics, the reproducibility of run experiments, as well as different levels of customization for different users. 

Fairlearn~\citep{werts2023}, and AIF360~\citep{bellamy2018} are popular fair ML packages  to facilitate adoption by offering methods~\citep{feldman2015,hardt2016,agarwal2018}, fairness metrics~\citep{hardt2016} and datasets available in the literature~\citep{kohavi1995,angwin2016,dua2019,ding2021}. However, some issues hinder their usability as standard toolkits for fairness studies. First, both lack a defined experimentation pipeline, requiring users to opt for external packages for fundamental tasks, such as dataset splitting and hyperparameter optimization~\citep{schelter2019}. Second, inconsistency in class behavior and implementation force users to customize the code depending on the  methods used. For instance, in AIF360's \texttt{DisparateImpactRemover} class, most of the parent class methods are not implemented. These issues create a high barrier for users to effectively use the packages. Our work tackles the lack of standardized tools for experimentation with fair ML, with an emphasis on the extensibility of methods, datasets, and metrics, the reproducibility of the experiments, as well as different levels of customization for different user needs.

%The two more popular packages for fair ML are Fairlearn~\citep{werts2023}, and AIF360~\citep{bellamy2018}. Both aim to facilitate adoption with an array of methods~\citep{feldman2015,hardt2016,agarwal2018}, common fairness metrics~\citep{hardt2016} and datasets used in the literature~\citep{kohavi1995,angwin2016,dua2019,ding2021}. However, these have specific issues that hinder their usability as standard toolkits for fairness studies in the ML community, namely the lack of a defined pipeline for experimentation and lack of consistency between classes of the same group. In this work, we propose to tackle the lack of standardized tools for experimentation with fair ML, with an emphasis on the extensibility of methods, datasets, and metrics, the reproducibility of run experiments, as well as different levels of customization for different users. 

% ----------------------------------------------------------
%
%     FRAMEWORK
%
% ----------------------------------------------------------

\section{Aequitas Flow}
\label{sec:Aequitas_flow}

\begin{figure}[t]
    \centering
    \noindent\makebox[\textwidth]{\includegraphics[width = 1.02\textwidth]{"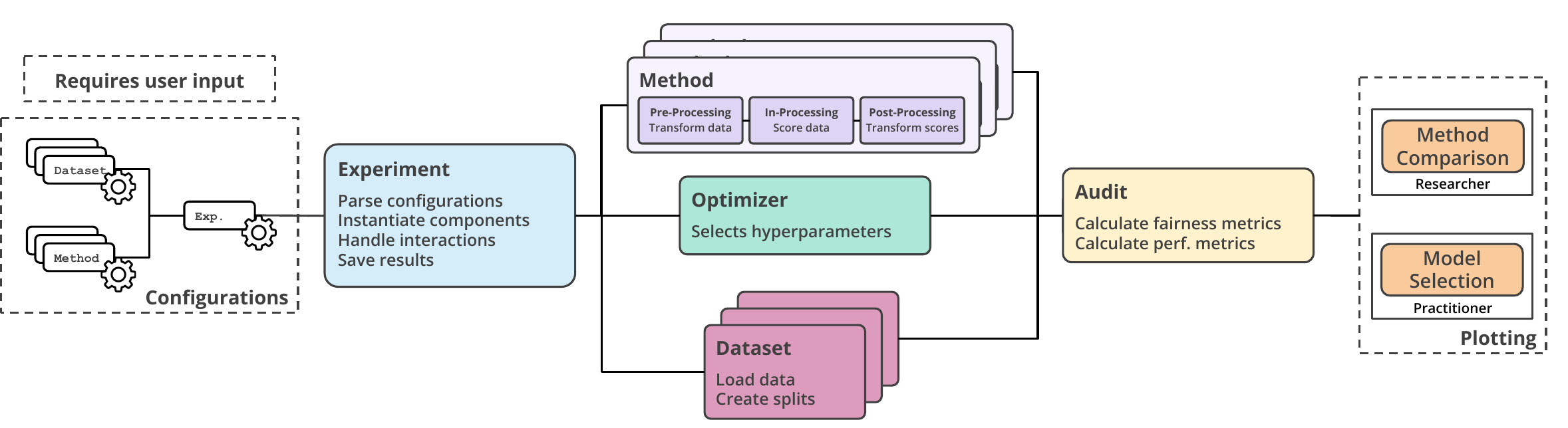"}}
    \caption{\textbf{Diagram of an Experiment in Aequitas Flow.} The user input is passed to the Experiment, which will instantiate the components (Methods, Datasets, and Optimizer) in the pipeline. For each target task (for a researcher or practitioner), different plotting methods can be used to analyze the experimental results.}
    \label{fig:diagram}
\end{figure}

The Aequitas Flow package is a comprehensive framework that integrates the necessary elements for a complete fair ML experiment. These are methods, datasets, and optimization strategies. They can be accessed through a standardized pipeline defined by configuration files, Python dictionaries or instantiated independently. This provides a standardized platform for experimental fairness testing. Figure \ref{fig:diagram} represents the pipeline's structure, mapping the components and interactions and offering an overview of the fairness experimentation process.

\paragraph{Experiment:}The \texttt{Experiment} is the main component that orchestrates the workflow within the package. It processes input configurations, which can be provided as either files\footnote{Examples provided in the repository.}  or Python dictionaries. These specify the methods, datasets, and optimization parameters. The Experiment component initializes and populates the necessary classes, ensuring they interact deterministically throughout the execution process. When an experiment is completed, the results can be analyzed directly within the class with the appropriate methods. A variant of this component allows for simplified usage as it only requires the definition of a dataset. This feature is designed to streamline initial experiments and reduce configurations.

% \begin{minted}[bgcolor=blue!13, fontsize=\small]{python}
\begin{lstlisting}
exp = Experiment(config_file="configs/experiment.yaml")
exp.run()
\end{lstlisting}    

\paragraph{Optimizer:}The \texttt{Optimizer} component manages hyperparameter selection and model evaluation. It receives the hyperparameter search space of the methods and a split dataset to conduct hyperparameter tuning. It evaluates the performance of models, and stores the resulting artifacts. The component uses Optuna~\citep{akiba2019} for hyperparameter selection and the bias auditing functionality of Aequitas~\citep{saleiro2018} for fairness and performance evaluation. This component should only be instantiated by an \texttt{Experiment}, to guarantee consistency in input arguments. Several attributes of the hyperparameter optimization can be determined by configurations, such as the number of trials and jobs, the selection algorithm (\textit{e.g.}, random search, grid search), and the random seed.

\paragraph{Datasets:} This component has two primary functions: loading the data and generating splits. It maintains information about the prediction target, typing, and sensitive features. The data is stored in a pandas dataframe format~\citep{reback2020pandas}. The framework initially encompasses eleven tabular datasets, including those from the BankAccountFraud~\citep{jesus2022} and Folktables~\citep{ding2021}.  The component also permits user-supplied datasets in CSV or parquet formats with splits based on a column, or randomly.

%\begin{minted}[bgcolor=blue!13, fontsize=\small]{python}
\begin{lstlisting} 
dataset = datasets.FolkTables(variant='ACSIncome')
dataset.load_data()
dataset.create_splits()
dataset.train.X  # return the train feature matrix
\end{lstlisting}

%with either random splits or based on a column value. 

\paragraph{Methods:}This group of components handles data processing and creates and adjusts predictions for validation and test sets. Aequitas Flow provides interfaces for the three recognized types of fair ML methods~\citep{caton2020,mehrabi2021,pessach2022}: pre-processing, in-processing, and post-processing. Pre-processing methods modify the input data, in-processing methods typically directly modify the objective function and generate prediction scores, and post-processing methods adjust these scores or rankings. Additionally,  ML classification methods are included in the category of base estimators and function similarly to in-processing methods. The methods adhere to a standardized interface to facilitate calls within the experiment class. In the current version of Aequitas, 15 methods are supported. 
% Aequitas Flow includes pre-processing techniques such as undersampling and oversampling~\citep{lamba2021}, label massaging and supression~\citep{kamiran2011}, and label flipping~\citep{silva2023}. For in-processing, the package includes FairGBM~\citep{cruz2023}, a gradient boosting machine algorithm with fairness constraints, and the methods of Exponentiated Gradient and Grid Search~\citep{agarwal2018,werts2023} for transforming fairness constraints into cost-sensitive classification. For post-processing, the framework includes group-wise thresholding~\citep{hardt2016}.

% \begin{minted}[bgcolor=blue!13, fontsize=\small]{python}
\begin{lstlisting}
model = methods.inprocessing.FairGBM()
model.fit(train.X, train.y, train.s)
preds = model.predict_proba(val.X, val.s)
\end{lstlisting}

\paragraph{Audit:}The Aequitas toolkit offers a suite of metrics based on the confusion matrix for the protected groups in the dataset. Users may specify a group as a reference for comparison and select the appropriate fairness metric for their analysis. Experiments leverage the \texttt{Audit} class to calculate metrics and disparities when analyzing the produced prediction scores of a model.

% \begin{minted}[bgcolor=blue!13, fontsize=\small]{python}
\begin{lstlisting}
audit_df = pd.DataFrame({"score": preds, "label": val.y, "group": val.s})
audit = Audit(audit_df)
audit.performance()  # Obtain performance metrics
audit.audit()  # Obtain fairness metrics
\end{lstlisting}

\begin{figure}[h]
\begin{tabular}{cc}
  \includegraphics[width=0.44\linewidth]{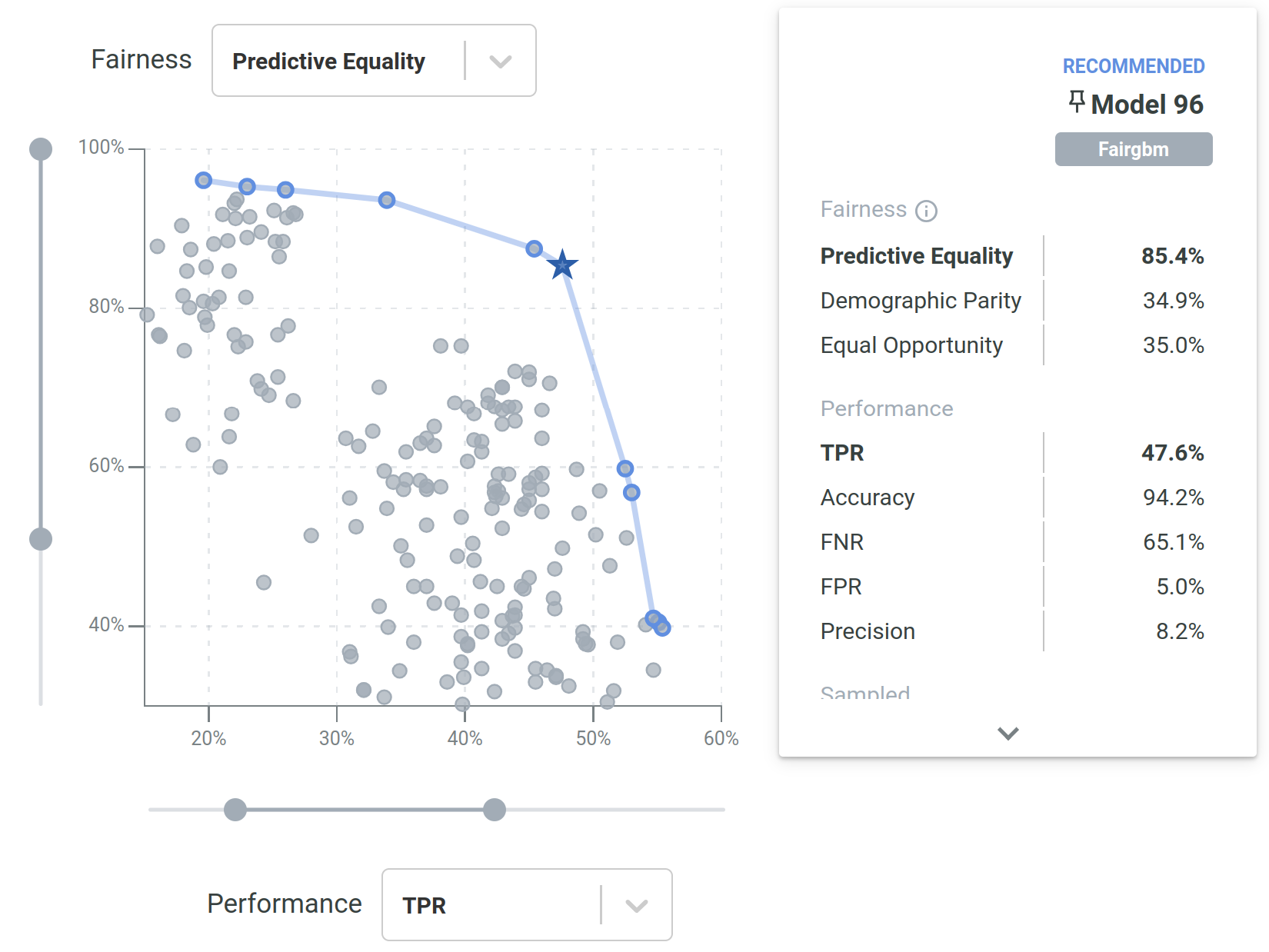} &   \includegraphics[width=0.48\linewidth]{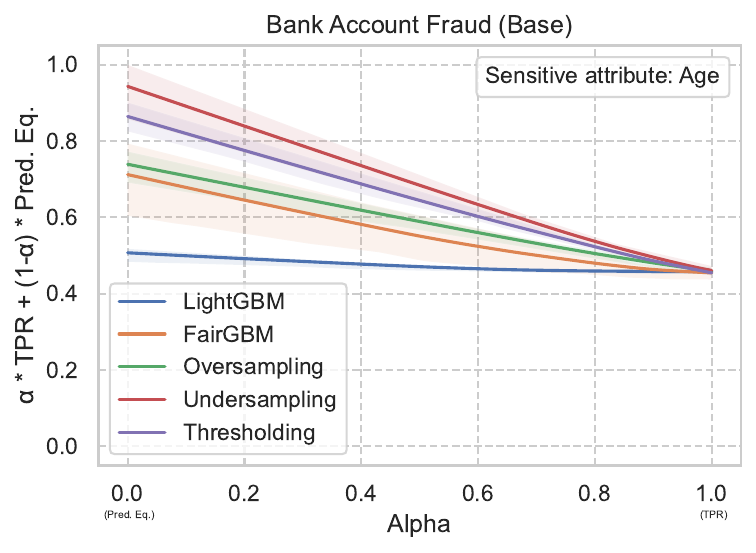} \\
(a) & (b) \\[6pt]
%\includegraphics[width=0.5\linewidth]{figures/summary_chart.pdf} &   \includegraphics[width=0.5\linewidth]{figures/disparity_chart.pdf} \\
%(c)  & (d)\\
\end{tabular}
\caption{Plots introduced in Aequitas Flow. Plot (a) is designed for model selection; Plot (b) compares the different tested methods.}
\label{fig:plots}
\end{figure}

\paragraph{Plotting:} Aequitas Flow provides two workflows based on the goal of the user. The first is around model selection (a), where users can plot the trained models with the desired metrics of fairness and performance in each axis. The Pareto frontier is displayed, with the model with the best fairness-performance trade-off highlighted. The second provides a comparison of methods (b). Confidence intervals for the combined performance and fairness are calculated for each tested method in the trade-offs of these metrics. Additional plotting methods are available for in-depth bias auditing. Figure \ref{fig:plots} shows examples of both.

\section{Conclusion}

Aequitas Flow is an open-source framework that makes end-to-end experimentation with fair ML  easier through the use of customizable components, namely datasets, methods, metrics, and optimization algorithms. It enhances robustness and reproducibility by addressing the issues of ad-hoc and single-use setups in fair ML experimentation. This can  lead to better benchmarking and adoption of fair ML techniques in real world settings. While initially focused on tabular datasets, the framework's flexible interfaces allow adaptation to other data formats, and ongoing updates will incorporate additional implementations, in a welcoming environment to community contributions. Recognizing the challenges associated with responsibly using this framework in real-world applications, we aim to support the widespread adoption of fair ML methodologies and increase their societal impact.

\bibliography{refs}

\begin{thebibliography}{32}
\providecommand{\natexlab}[1]{#1}
\providecommand{\url}[1]{\texttt{#1}}
\expandafter\ifx\csname urlstyle\endcsname\relax
  \providecommand{\doi}[1]{doi: #1}\else
  \providecommand{\doi}{doi: \begingroup \urlstyle{rm}\Url}\fi

\bibitem[Agarwal et~al.(2018)Agarwal, Beygelzimer, Dudik, Langford, and
  Wallach]{agarwal2018}
Alekh Agarwal, Alina Beygelzimer, Miroslav Dudik, John Langford, and Hanna
  Wallach.
\newblock A reductions approach to fair classification.
\newblock In Jennifer Dy and Andreas Krause, editors, \emph{Proceedings of the
  35th International Conference on Machine Learning}, volume~80 of
  \emph{Proceedings of Machine Learning Research}, pages 60--69. PMLR, 10--15
  Jul 2018.
\newblock URL \url{https://proceedings.mlr.press/v80/agarwal18a.html}.

\bibitem[Akiba et~al.(2019)Akiba, Sano, Yanase, Ohta, and Koyama]{akiba2019}
Takuya Akiba, Shotaro Sano, Toshihiko Yanase, Takeru Ohta, and Masanori Koyama.
\newblock Optuna: A next-generation hyperparameter optimization framework.
\newblock In \emph{Proceedings of the 25th ACM SIGKDD International Conference
  on Knowledge Discovery \& Data Mining}, KDD '19, page 2623–2631, New York,
  NY, USA, 2019. Association for Computing Machinery.
\newblock ISBN 9781450362016.
\newblock \doi{10.1145/3292500.3330701}.
\newblock URL \url{https://doi.org/10.1145/3292500.3330701}.

\bibitem[Angwin et~al.(2016)Angwin, Larson, Mattu, and Kirchner]{angwin2016}
Julia Angwin, Jeff Larson, Surya Mattu, and Lauren Kirchner.
\newblock Machine bias.
\newblock In \emph{Ethics of data and analytics}, pages 254--264. Auerbach
  Publications, 2016.

\bibitem[Bartlett et~al.(2019)Bartlett, Morse, Stanton, and
  Wallace]{bartlett2019}
Robert Bartlett, Adair Morse, Richard Stanton, and Nancy Wallace.
\newblock {Consumer-Lending Discrimination in the FinTech Era}.
\newblock Technical report, National Bureau of Economic Research, 2019.

\bibitem[Bellamy et~al.(2018)Bellamy, Dey, Hind, Hoffman, Houde, Kannan, Lohia,
  Martino, Mehta, Mojsilovic, Nagar, Ramamurthy, Richards, Saha, Sattigeri,
  Singh, Varshney, and Zhang]{bellamy2018}
Rachel K.~E. Bellamy, Kuntal Dey, Michael Hind, Samuel~C. Hoffman, Stephanie
  Houde, Kalapriya Kannan, Pranay Lohia, Jacquelyn Martino, Sameep Mehta,
  Aleksandra Mojsilovic, Seema Nagar, Karthikeyan~Natesan Ramamurthy, John
  Richards, Diptikalyan Saha, Prasanna Sattigeri, Moninder Singh, Kush~R.
  Varshney, and Yunfeng Zhang.
\newblock {AI Fairness} 360: An extensible toolkit for detecting,
  understanding, and mitigating unwanted algorithmic bias, October 2018.
\newblock URL \url{https://arxiv.org/abs/1810.01943}.

\bibitem[Calders and Verwer(2010)]{calders2010}
Toon Calders and Sicco Verwer.
\newblock Three naive bayes approaches for discrimination-free classification.
\newblock \emph{Data Mining and Knowledge Discovery}, 21\penalty0 (2):\penalty0
  277--292, Sep 2010.
\newblock ISSN 1573-756X.
\newblock \doi{10.1007/s10618-010-0190-x}.
\newblock URL \url{https://doi.org/10.1007/s10618-010-0190-x}.

\bibitem[Calmon et~al.(2017)Calmon, Wei, Vinzamuri, Ramamurthy, and
  Varshney]{calmon2017}
Flavio~P. Calmon, Dennis Wei, Bhanukiran Vinzamuri, Karthikeyan~Natesan
  Ramamurthy, and Kush~R. Varshney.
\newblock Optimized pre-processing for discrimination prevention.
\newblock In \emph{Proceedings of the 31st International Conference on Neural
  Information Processing Systems}, NIPS'17, page 3995–4004, Red Hook, NY,
  USA, 2017. Curran Associates Inc.
\newblock ISBN 9781510860964.

\bibitem[Caton and Haas(2023)]{caton2020}
Simon Caton and Christian Haas.
\newblock Fairness in machine learning: A survey.
\newblock \emph{ACM Comput. Surv.}, aug 2023.
\newblock ISSN 0360-0300.
\newblock \doi{10.1145/3616865}.
\newblock URL \url{https://doi.org/10.1145/3616865}.

\bibitem[Chouldechova(2017)]{chouldechova2017}
Alexandra Chouldechova.
\newblock {Fair Prediction with Disparate Impact: A Study of Bias in Recidivism
  Prediction Instruments}.
\newblock \emph{Big Data}, 5\penalty0 (2):\penalty0 153--163, jun 2017.
\newblock ISSN 2167-6461.
\newblock \doi{10.1089/big.2016.0047}.
\newblock URL \url{http://www.liebertpub.com/doi/10.1089/big.2016.0047}.

\bibitem[Corbett-Davies et~al.(2017)Corbett-Davies, Pierson, Feller, Goel, and
  Huq]{davies2017}
Sam Corbett-Davies, Emma Pierson, Avi Feller, Sharad Goel, and Aziz Huq.
\newblock Algorithmic decision making and the cost of fairness.
\newblock In \emph{Proceedings of the 23rd ACM SIGKDD International Conference
  on Knowledge Discovery and Data Mining}, KDD '17, page 797–806, New York,
  NY, USA, 2017. Association for Computing Machinery.
\newblock ISBN 9781450348874.
\newblock \doi{10.1145/3097983.3098095}.
\newblock URL \url{https://doi.org/10.1145/3097983.3098095}.

\bibitem[Cotter et~al.(2019)Cotter, Jiang, and Sridharan]{cotter2018}
Andrew Cotter, Heinrich Jiang, and Karthik Sridharan.
\newblock Two-player games for efficient non-convex constrained optimization.
\newblock In Aurélien Garivier and Satyen Kale, editors, \emph{Proceedings of
  the 30th International Conference on Algorithmic Learning Theory}, volume~98
  of \emph{Proceedings of Machine Learning Research}, pages 300--332. PMLR,
  22--24 Mar 2019.
\newblock URL \url{https://proceedings.mlr.press/v98/cotter19a.html}.

\bibitem[Dastin(2018)]{dastin2018}
Jeffrey Dastin.
\newblock Amazon scraps secret ai recruiting tool that showed bias against
  women.
\newblock In \emph{Ethics of data and analytics}, pages 296--299. Auerbach
  Publications, 2018.

\bibitem[Ding et~al.(2021)Ding, Hardt, Miller, and Schmidt]{ding2021}
Frances Ding, Moritz Hardt, John Miller, and Ludwig Schmidt.
\newblock Retiring adult: New datasets for fair machine learning.
\newblock In M.~Ranzato, A.~Beygelzimer, Y.~Dauphin, P.S. Liang, and J.~Wortman
  Vaughan, editors, \emph{Advances in Neural Information Processing Systems},
  volume~34, pages 6478--6490. Curran Associates, Inc., 2021.
\newblock URL
  \url{https://proceedings.neurips.cc/paper_files/paper/2021/file/32e54441e6382a7fbacbbbaf3c450059-Paper.pdf}.

\bibitem[Dua and Graff(2017)]{dua2019}
Dheeru Dua and Casey Graff.
\newblock {UCI} machine learning repository, 2017.
\newblock URL \url{http://archive.ics.uci.edu/ml}.

\bibitem[Dwork et~al.(2012)Dwork, Hardt, Pitassi, Reingold, and
  Zemel]{dwork2012}
Cynthia Dwork, Moritz Hardt, Toniann Pitassi, Omer Reingold, and Richard Zemel.
\newblock {Fairness through awareness}.
\newblock In \emph{Proc. of the 3rd Innovations in Theoretical Computer Science
  Conf. on - ITCS '12}, pages 214--226, New York, USA, 2012. ACM Press.
\newblock ISBN 9781450311151.
\newblock \doi{10.1145/2090236.2090255}.
\newblock URL \url{http://arxiv.org/abs/1104.3913}.

\bibitem[Feldman et~al.(2015)Feldman, Friedler, Moeller, Scheidegger, and
  Venkatasubramanian]{feldman2015}
Michael Feldman, Sorelle~A. Friedler, John Moeller, Carlos Scheidegger, and
  Suresh Venkatasubramanian.
\newblock Certifying and removing disparate impact.
\newblock In \emph{Proceedings of the 21th ACM SIGKDD International Conference
  on Knowledge Discovery and Data Mining}, KDD '15, page 259–268, New York,
  NY, USA, 2015. Association for Computing Machinery.
\newblock ISBN 9781450336642.
\newblock \doi{10.1145/2783258.2783311}.
\newblock URL \url{https://doi.org/10.1145/2783258.2783311}.

\bibitem[Fish et~al.(2016)Fish, Kun, and Lelkes]{fish2016}
Benjamin Fish, Jeremy Kun, and {\'{A}}d{\'{a}}m~D{\'{a}}niel Lelkes.
\newblock A confidence-based approach for balancing fairness and accuracy.
\newblock In Sanjay~Chawla Venkatasubramanian and Wagner~Meira Jr., editors,
  \emph{Proceedings of the 2016 {SIAM} International Conference on Data Mining,
  Miami, Florida, USA, May 5-7, 2016}, pages 144--152. {SIAM}, 2016.
\newblock \doi{10.1137/1.9781611974348.17}.
\newblock URL \url{https://doi.org/10.1137/1.9781611974348.17}.

\bibitem[Friedler et~al.(2019)Friedler, Scheidegger, Venkatasubramanian,
  Choudhary, Hamilton, and Roth]{friedler2019}
Sorelle~A. Friedler, Carlos Scheidegger, Suresh Venkatasubramanian, Sonam
  Choudhary, Evan~P. Hamilton, and Derek Roth.
\newblock A comparative study of fairness-enhancing interventions in machine
  learning.
\newblock In \emph{Proceedings of the Conference on Fairness, Accountability,
  and Transparency}, FAT* '19, page 329–338, New York, NY, USA, 2019.
  Association for Computing Machinery.
\newblock ISBN 9781450361255.
\newblock \doi{10.1145/3287560.3287589}.
\newblock URL \url{https://doi.org/10.1145/3287560.3287589}.

\bibitem[Hardt et~al.(2016)Hardt, Price, and Srebro]{hardt2016}
Moritz Hardt, Eric Price, and Nathan Srebro.
\newblock Equality of opportunity in supervised learning.
\newblock In \emph{Proceedings of the 30th International Conference on Neural
  Information Processing Systems}, NIPS'16, page 3323–3331, Red Hook, NY,
  USA, 2016. Curran Associates Inc.
\newblock ISBN 9781510838819.

\bibitem[Igoe(2021)]{igoe2021}
K~Igoe.
\newblock Algorithmic bias in health care exacerbates social inequities--how to
  prevent it.
\newblock \emph{Executive and Continuing Professional Education}, 2021.

\bibitem[Jesus et~al.(2022)Jesus, Pombal, Alves, Cruz, Saleiro, Ribeiro, Gama,
  and Bizarro]{jesus2022}
S{\'{e}}rgio Jesus, Jos{\'{e}} Pombal, Duarte Alves, Andr{\'{e}}~Ferreira Cruz,
  Pedro Saleiro, Rita~P. Ribeiro, Jo{\~{a}}o Gama, and Pedro Bizarro.
\newblock Turning the tables: Biased, imbalanced, dynamic tabular datasets for
  {ML} evaluation.
\newblock In \emph{NeurIPS}, 2022.
\newblock URL
  \url{http://papers.nips.cc/paper\_files/paper/2022/hash/d9696563856bd350e4e7ac5e5812f23c-Abstract-Datasets\_and\_Benchmarks.html}.

\bibitem[Kohavi(1996)]{kohavi1995}
Ron Kohavi.
\newblock Scaling up the accuracy of naive-bayes classifiers: A decision-tree
  hybrid.
\newblock In \emph{Proceedings of the Second International Conference on
  Knowledge Discovery and Data Mining}, KDD'96, page 202–207. AAAI Press,
  1996.

\bibitem[Lamba et~al.(2021)Lamba, Rodolfa, and Ghani]{lamba2021}
Hemank Lamba, Kit Rodolfa, and Rayid Ghani.
\newblock An empirical comparison of bias reduction methods on real-world
  problems in high-stakes policy settings.
\newblock \emph{ACM SIGKDD Explorations Newsletter}, 23:\penalty0 69--85, 05
  2021.
\newblock \doi{10.1145/3468507.3468518}.

\bibitem[Lee and Singh(2021)]{lee2021}
Michelle Seng~Ah Lee and Jat Singh.
\newblock The landscape and gaps in open source fairness toolkits.
\newblock In \emph{Proceedings of the 2021 CHI Conference on Human Factors in
  Computing Systems}, CHI '21, New York, NY, USA, 2021. Association for
  Computing Machinery.
\newblock ISBN 9781450380966.
\newblock \doi{10.1145/3411764.3445261}.
\newblock URL \url{https://doi.org/10.1145/3411764.3445261}.

\bibitem[Mehrabi et~al.(2021)Mehrabi, Morstatter, Saxena, Lerman, and
  Galstyan]{mehrabi2021}
Ninareh Mehrabi, Fred Morstatter, Nripsuta Saxena, Kristina Lerman, and Aram
  Galstyan.
\newblock A survey on bias and fairness in machine learning.
\newblock \emph{ACM Comput. Surv.}, 54\penalty0 (6), jul 2021.
\newblock ISSN 0360-0300.
\newblock \doi{10.1145/3457607}.
\newblock URL \url{https://doi.org/10.1145/3457607}.

\bibitem[pandas~development team(2020)]{reback2020pandas}
The pandas~development team.
\newblock pandas-dev/pandas: Pandas, February 2020.
\newblock URL \url{https://doi.org/10.5281/zenodo.3509134}.

\bibitem[Pessach and Shmueli(2022)]{pessach2022}
Dana Pessach and Erez Shmueli.
\newblock A review on fairness in machine learning.
\newblock \emph{ACM Comput. Surv.}, 55\penalty0 (3), feb 2022.
\newblock ISSN 0360-0300.
\newblock \doi{10.1145/3494672}.
\newblock URL \url{https://doi.org/10.1145/3494672}.

\bibitem[Saleiro et~al.(2018)Saleiro, Kuester, Stevens, Anisfeld, Hinkson,
  London, and Ghani]{saleiro2018}
Pedro Saleiro, Benedict Kuester, Abby Stevens, Ari Anisfeld, Loren Hinkson,
  Jesse London, and Rayid Ghani.
\newblock Aequitas: A bias and fairness audit toolkit.
\newblock \emph{arXiv preprint arXiv:1811.05577}, 2018.

\bibitem[Schelter et~al.(2019)Schelter, He, Khilnani, and
  Stoyanovich]{schelter2019}
Sebastian Schelter, Yuxuan He, Jatin Khilnani, and Julia Stoyanovich.
\newblock Fairprep: Promoting data to a first-class citizen in studies on
  fairness-enhancing interventions.
\newblock \emph{ArXiv}, abs/1911.12587, 2019.
\newblock URL \url{https://api.semanticscholar.org/CorpusID:208512964}.

\bibitem[Weerts et~al.(2023)Weerts, Dudík, Edgar, Jalali, Lutz, and
  Madaio]{werts2023}
Hilde Weerts, Miroslav Dudík, Richard Edgar, Adrin Jalali, Roman Lutz, and
  Michael Madaio.
\newblock Fairlearn: Assessing and improving fairness of ai systems, 2023.
\newblock URL \url{http://jmlr.org/papers/v24/23-0389.html}.

\bibitem[Zafar et~al.(2017)Zafar, Valera, Rogriguez, and Gummadi]{zafar2017}
Muhammad~Bilal Zafar, Isabel Valera, Manuel~Gomez Rogriguez, and Krishna~P.
  Gummadi.
\newblock {Fairness Constraints: Mechanisms for Fair Classification}.
\newblock In Aarti Singh and Jerry Zhu, editors, \emph{Proceedings of the 20th
  International Conference on Artificial Intelligence and Statistics},
  volume~54 of \emph{Proceedings of Machine Learning Research}, pages 962--970.
  PMLR, 20--22 Apr 2017.
\newblock URL \url{https://proceedings.mlr.press/v54/zafar17a.html}.

\bibitem[Zhang and Zhou(2019)]{zhang2019}
Yukun Zhang and Longsheng Zhou.
\newblock Fairness assessment for artificial intelligence in financial
  industry.
\newblock \emph{arXiv preprint arXiv:1912.07211}, 2019.

\end{thebibliography}
\end{document}